\documentclass{article}

\usepackage[preprint]{timeseries_workshop}

\usepackage[utf8]{inputenc} 
\usepackage[T1]{fontenc}    
\usepackage{url}            
\usepackage{booktabs}       
\usepackage{amsfonts}       
\usepackage{nicefrac}       
\usepackage{microtype}      
\usepackage{xcolor}         
\usepackage{graphicx}
\usepackage{multirow} 
\usepackage{float}
\usepackage{multirow}
\usepackage{amsmath}
\usepackage{hyperref}       
\usepackage{mymath}

\def\R{{\mathbb{R}}}

\title{Leveraging Generic Time Series Foundation Models for EEG Classification}

\author{%
  Th\'eo Gnassounou\thanks{equal contribution}\\
  Université Paris-Saclay, Inria, CEA\\ \And
  Yessin Moakher$^{*}$\\ Huawei Noah's Ark Lab \And Shifeng Xie \\ Huawei Noah's Ark Lab \And Vasilii Feofanov \\ Huawei Noah's Ark Lab \And Ievgen Redko \\ Huawei Noah's Ark Lab
}

\begin{document}

\maketitle

\begin{abstract}
Foundation models for time series are emerging as powerful general-purpose backbones, yet their potential for domain-specific biomedical signals such as electroencephalography (EEG) remains rather unexplored. In this work, we investigate the applicability a recently proposed time series classification foundation model, to a different EEG tasks such as motor imagery classification and sleep stage prediction. We test two pretraining regimes: (a) pretraining on heterogeneous real-world time series from multiple domains, and (b) pretraining on purely synthetic data. We find that both variants yield strong performance, consistently outperforming EEGNet, a widely used convolutional baseline, and CBraMod, the most recent EEG-specific foundation model. These results suggest that generalist time series foundation models, even when pretrained on data of non-neural origin or on synthetic signals, can transfer effectively to EEG. Our findings highlight the promise of leveraging cross-domain pretrained models for brain signal analysis, suggesting that EEG may benefit from advances in the broader time series literature.
\end{abstract}

\section{Introduction}

Electroencephalography (EEG) is a widely used non-invasive technique for monitoring brain activity, with applications ranging from clinical diagnostics to brain–computer interfaces (BCI). A central challenge in EEG analysis is classification, which underpins tasks such as motor imagery for BCI control~\citep{barachant2012bci}, sleep staging~\citep{chambon2018deep}, and emotion recognition~\citep{Li_2022}. Despite their promise, these applications face significant barriers: EEG datasets are typically small, fragmented across institutions, and/or difficult to share due to privacy concerns.
Furthermore, EEG signals exhibit strong variability across subjects and sessions, which makes generalization to unseen individuals especially difficult. This scarcity and variability limit the effectiveness of deep learning models such as CNNs~\citep{lawhern2018eegnet,chambon2018deep}, LSTMs~\citep{phan_seqsleepnet_2019}, and transformers~\citep{phan2022sleeptransformer,Guo2024transformer}.

In parallel, machine learning has been transformed by the rise of foundation models \citep{bommasani2021opportunities}. In computer vision \citep{dosovitskiy2021vit} and natural language processing~\citep{achiam2023gpt}, large-scale pretraining on diverse data has enabled models to generalize across tasks, reducing the need for task-specific architectures and large labeled datasets. Inspired by this success, time series foundation models (TSFMs) have recently emerged. Some focus on forecasting~\citep{ansari2024chronos,auer2025tirex}, others on classification~\citep{feofanov2025mantis}, also with some attempts to unify multiple time series tasks~\citep{goswami2024moment}. Interestingly, both real-world~\citep{feofanov2025mantis} and synthetic data~\citep{xie2025cauker} have been shown effective for pretraining such models.

In EEG specifically, efforts to build foundation models are more recent. CBraMod \citep{wang2025cbramod} introduced a masked-reconstruction approach pretrained on the large-scale TUEG corpus \citep{obeid2016TUEG}, showing encouraging results for different BCI tasks. Yet, its evaluation remains limited in scope, and questions persist about whether EEG-specific pretraining is necessary, or whether general-purpose TSFMs can transfer effectively to EEG.

This paper takes a step forward to address the aforementioned questions. We investigate the applicability of Mantis \citep{feofanov2025mantis}, the most recent time series classification foundation model pretrained either on heterogeneous time series datasets or synthetic data, to EEG signals. 
Across benchmarks for sleep staging and motor imagery classification, we find that Mantis achieves strong transfer performance, generally outperforming both EEGNet~\citep{lawhern2018eegnet}, a widely used baseline, and CBraMod, the most recent EEG-specific foundation model. This result, at the same time, gives a high promise on developing  general-purpose TSFMs and highlights a large room for improvement in brain signal analysis.
This finding suggests opportunities for leveraging cross-domain TSFMs in brain signal analysis as well as reveals current limitations of EEG-specific foundation models.

\section{Methodology}

\begin{table}[t]
    \centering
    \caption{Model comparison.}
    \resizebox{0.8\textwidth}{!}{\begin{tabular}{c|c c c c c}
        \toprule
         Model & Type & Domain & Size & Pretraining & Multivariate\\
          &  &  &  & Type & Pretraining \\
         \midrule
         EEGNet & Tailored & EEG & <0.01M & \texttt{NaN} & \texttt{NaN} \\
         CBraMod & Foundation Model & EEG & 4M & Reconstruction & Yes \\
         Mantis & Foundation Model & Generic & 8M & Contrastive & No \\
         \bottomrule
    \end{tabular}}
    \label{tab:model-comparison}
\end{table}

Time series classification foundation model is an encoder $F: \R^{C\times T} \to \R^{Q}$ that projects any signal $\mbf{x}\in\R^{C\times T}$ with $C$ channels and sequence length $T$ to a discriminative hidden space $\R^{Q}$. During pretraining, we observe an unlabeled pretraining set $\mathrm{X}_{\text{0}}$ that is sufficiently large for learning rich embeddings 
that generalize well across different tasks. During fine-tuning, we observe a supervised downstream task with observations $\{\mbf{x}_i, y_i\}_{i=1}^n$. 
We append a classification head $h:\R^{Q}\to \R^K$ and fine-tune $h\circo F$ by minimizing the cross-entropy loss. In this work, we consider two foundation models, which we briefly present below, and the summary can be found in Table \ref{tab:model-comparison}. 

The first foundation model is CBraMod~\citep{wang2025cbramod}, recently proposed for EEG data. It is a masked autoencoder~\citep{he2022masked} focused on correct reconstruction of missing patches. The model has been pretrained on the TUEG dataset \citep{obeid2016TUEG} with 1,109,545 EEG samples after pre-processing. One of the main features of the model is that it is pretrained directly on the multivariate signals. This is achieved by the proposed criss-cross transformer that mixes time-wise and channel-wise attention modules. More implementation details can be found in Appendix \ref{sec:cbramod}.

Second, we consider Mantis \citep{feofanov2025mantis}, a foundation model designed for general-purpose time series classification. In contrast to CBraMod, Mantis is pretrained using contrastive learning, pushing different augmentations of a single time series to lie close in the representation space.
Originally, they have pretrained Mantis on a mix of different real-world time series datasets (1.8 millions samples in total, \citeauthor{lin2024nutime},\citeyear{lin2024nutime}), with a small portion of EEG data. Recently, \citet{xie2025cauker} has shown that Mantis achieves the same performance by being pretrained on a purely synthetic dataset generated by the CauKer algorithm (1 million samples). In our experiment, we will consider both these checkpoints. We give more details in Appendix \ref{sec:mantis}.

It is worth mentioning that in our preliminary experiments, we have found that freezing the encoder for EEG data leads to a huge decrease in performance, so fine-tuning is necessary in this context. This is why we have not considered MOMENT \citep{goswami2024moment}, which is very difficult to fine-tune due to its large model size compared to CBraMod and Mantis. In our experiments, we compare the two foundation models with  EEGNet~\citep{lawhern2018eegnet}, a classical CNN architecture specifically designed for EEG signals.

\section{Experimental Results}
\label{sec:exp-res}

\subsection{Setup}
We conduct two different sets of experiments to evaluate Mantis on EEG data. First, we follow the experimental setup used in CBraMod~\citep{wang2025cbramod} and concentrate on motor imagery classification. Second, we perform a comprehensive study on 8 sleep stage prediction datasets following \citet{Perslev2021Usleep} and \citet{gnassounou2025psdnorm}. While in the first case we extract the results of CBraMod and EEGNet from \citet{wang2025cbramod}, in the second case, we fine-tune these models, so we can test the adaptability of CBraMod to new EEG tasks. 

\paragraph{BCI dataset}
In the first experiments we use two dataset of Brain Computer Interface for Motor Imagery classification. PhysioNet-MI~\citep{schalk2004PhysiontMI} comprises 109 subjects with 64 channels with a sampling rate of 160 Hz. This dataset includes four different motor imagery classes: left hand, right hand, both hands, and both feet. SHU-MI~\citep{ma2022shumi} comprises 25 subjects with 32 EEG channels sample at 250Hz. This dataset covers binary motor imagery with the right hand and the left hand. 
Each dataset is resampled at 200 Hz. For PhysioNet-MI, subjects 1–70, 71–89, and 90–109 are used for training, validation, and testing, respectively. For SHU-MI, subjects 1-15, 16-20, 21-25 for training, validation, and testing, respectively.

\paragraph{Sleep datasets}

In the experiments, we used 8 sleep staging datasets.
ABC~\citep{jp2018ABC}, CCSHS~\citep{rosen_prevalence_2003ccshs}, 
CFS~\citep{redline_familial_1995cfs}, HPAP~\citep{rosen2012homepap}, MROS~\citep{blackwell_associations_2011Mros}, SHHS~\citep{SHHS}, CHAT~\citep{marcus2013CHAT}, and SOF~\citep{spira_sleep-disordered_2008SOF}
are publicly available sleep datasets with restricted access
from National Sleep Research Resource (NSRR)~\citep{zhang_national_2018}.
PhysioNet~\citep{Phys} and MASS~\citep{MASS} are two other datasets publicly available. 
These datasets are recordings of one night of sleep of different patients.
Every 30\,s epoch of sleep is labeled with one of the five sleep stages: Wake, N1, N2, N3, and REM.
Datasets are split by subjects into training, validation, and test sets (60\%/20\%/20\%).
More details about the pre-processing is given in the Appendix.

\paragraph{Architecture setup}
For fine-tuning, Mantis use a linear layer with pre-LayerNorm as a classfication head. In the CBraMod's paper, they tune the head for each task, while in our experiments, we fixed it as a 3-layer MLP with ELU and dropout.
For the BCI experiments, models were trained for a maximum of 20 epochs using the AdamW optimizer with a batch size of 64 and a
weight decay of 0.01. We set the initial learning rate to  $1\times10^{-4}$ for the Physionet-MI dataset and $5\times10^{-4}$ for the SHU-MI
dataset. The learning rate was managed by a cosine scheduler with a warmup period over the first 20\% of training steps. We applied gradient clipping at a norm of 1.0 and utilized an early stopping mechanism with a patience of 3 epochs to prevent overfitting.

For sleep staging, models were trained for a maximum of 50 epochs using the AdamW optimizer with a batch size of 64 and a weight decay of 0.01. Training is monitored with early stopping on the validation set, using a patience of 5 epochs.
The learning rate is set to $1\times10^{-4}$ for Mantis and CBraMod, and $1\times10^{-3}$ for EEGNet.
To account for limited resources, we impose a maximum training time of 5 hours. A value of NaN indicates that the time limit was reached.

\subsection{Results on Brain Computer Interface}
We evaluated Mantis on a BCI motor imagery task, using the CBraMod setup for comparison (Table~\ref{tab:motor-imagery-results}). On the Physionet-MI dataset, Mantis achieves performance competitive with CBraMod, a specialized model for EEG that already surpasses classical CNNs like EEGNet by  6\%. This similar result is particularly noteworthy as Mantis was pretrained with minimal EEG data. More importantly, Mantis significantly outperforms the baseline on the SHU-MI dataset, achieving a 72.15\% F1-score versus CBraMod's 69.88\%, when Mantis was pretrained only on synthetic data \citep{xie2025cauker}. These results demonstrate that its architecture can achieve state-of-the-art performance on brain signals without extensive domain-specific pretraining.

Mantis's performance is even more compelling given its approach to modeling spatial correlations, which are critical for BCI tasks~\citep{barachant2012bci}. Unlike CBraMod, which relies on multivariate pretraining, Mantis processes channels univariately and only models their inter-dependencies at the final classification layer. The fact that Mantis still outperforms the multivariate approach suggests its architecture offers a more efficient method for preserving and leveraging spatial information in EEG signals.

\subsection{Results on Sleep Staging}

We then evaluate the models on sleep staging, a task characterized by a low number of EEG channels (typically 1-7)~\citep{Supratak_2017, gnassounou2024multisourcetesttimedomainadaptation, wang2025cbramod}, contrasting with the highly multivariate signals CBraMod was designed for. As shown in Table~\ref{tab:sleep_results} for a 2-channel setup, both foundation models surpass the EEGNet baseline. Crucially, Mantis consistently outperforms CBraMod across all pretraining configurations (real and synthetic data), with performance gains ranging from 0.3\% on CCSHS to nearly 3\% on the Mass dataset. This suggests that in scenarios with limited spatial information, CBraMod's multivariate architecture is less effective, whereas Mantis's more general, channel-independent design holds a distinct advantage.

Additionally, our results confirm the value of pretraining, as starting from a checkpoint yields better performance than random initialization. Pretraining also enhances training stability and efficiency, preventing runtime errors (denoted by NaN for runs exceeding 5 hours) that occurred when training from scratch. However, the performance gains are modest, indicating substantial room for improvement in future pretraining strategies for EEG data.

\begin{table}[ht]
    \centering
    
    \caption{\textbf{Three different scores for BCI over two datasets averaged over 3 seeds.} EEGNet and CBraMod results are from~\citep{wang2025cbramod}. 
    For Mantis, we report random initialization (Random), pretraining on real dataset (Real Pretrain) and synthetic pretraining on data generated by CauKer~\citep{xie2025cauker} (Synth Pretrain).
    }
    \label{tab:BCI_results}
    \resizebox{\textwidth}{!}{\begin{tabular}{llccccc}
    \toprule
    \multirow{2}{*}{Dataset}      & \multirow{2}{*}{Metric} & \multirow{2}{*}{EEGNet} & \multirow{2}{*}{CBraMod}    & \multicolumn{3}{c}{Mantis}                                                    \\
    \cmidrule(lr){5-7}
                                  & \multicolumn{1}{l}{}                        &                         &                             & Random             & Real Pretrain               & Synth Pretrain             \\
    \midrule
    \multirow{3}{*}{PhysioNet-MI} & Balanced Acc                                & $58.14_{\pm 1.25}$      & $64.17_{\pm 0.90}$          & $60.76_{\pm 1.10}$ & $\mathbf{64.43_{\pm 1.50}}$ & $61.90_{\pm 2.01}$         \\
                                  & Cohen’s Kappa                               & $44.68_{\pm 1.20}$      & $\mathbf{52.22_{\pm 1.70}}$ & $47.70_{\pm 1.46}$ & $52.13_{\pm 1.87}$          & $49.20_{\pm 3.35}$         \\
                                  & Weighted F1                                 & $57.96_{\pm 1.15}$      & $64.27_{\pm 1.00}$          & $60.44_{\pm 1.27}$ & $\mathbf{64.34_{\pm 1.63}}$ & $61.95_{\pm 2.52}$         \\
    \midrule
    \multirow{3}{*}{SHU-MI}       & Balanced Acc                                & $58.89_{\pm 1.77}$      & $63.70_{\pm 1.50}$          & $60.70_{\pm 1.90}$ & $63.00_{\pm 1.37}$          & $\mathbf{65.5_{\pm 4.3}}$  \\
                                  & AUROC                                       & $63.11_{\pm 1.42}$      & $\mathbf{71.39_{\pm 0.90}}$ & $68.10_{\pm 0.07}$ & $69.46_{\pm 1.18}$          & $70.90_{\pm 4.1}$          \\
                                  & AUC-PR                                      & $62.83_{\pm 1.52}$      & $69.88_{\pm 0.07}$          & $70.00_{\pm 0.09}$ & $70.55_{\pm 2.00}$          & $\mathbf{72.15_{\pm 3.8}}$\\
    \bottomrule
    \end{tabular}
    }
    \label{tab:motor-imagery-results}
\end{table}

\begin{table}[ht]
    \centering
    \caption{\textbf{Weighted F1 score for sleep staging over eight datasets averaged over 3 random seeds.} 
    Random and Synth Pretrain settings are as described in Table \ref{tab:BCI_results}.}
    \resizebox{\textwidth}{!}{\begin{tabular}{lcccccc}
    \toprule
    \multirow{2}{*}{Dataset}  & \multirow{2}{*}{EEGNet} & \multicolumn{2}{c}{CBraMod}         & \multicolumn{3}{c}{Mantis}                                               \\
    \cmidrule(lr){3-4} \cmidrule(lr){5-7}
     &                         & Random           & EEG Pretrain  & Random           & Real Pretrain          & Synth Pretrain         \\
    \midrule
    ABC        & $67.94_{\pm 6.52}$        & $70.61_{\pm 3.29}$ & $74.90_{\pm 4.89}$ & $72.82_{\pm 3.89}$ & $75.50_{\pm 5.62}$ & $\mathbf{75.74_{\pm 4.32}}$ \\
     CCSHS      & $83.13_{\pm 0.10}$        & $87.01_{\pm 0.27}$ & $88.04_{\pm 0.59}$ & $88.55_{\pm 0.39}$ & $\mathbf{88.85_{\pm 0.48}}$ & $88.80_{\pm 0.30}$ \\
     CFS        & $78.60_{\pm 1.31}$        & $83.48_{\pm 0.23}$ & $84.30_{\pm 0.08}$ & $84.96_{\pm 0.43}$ & $\mathbf{85.35_{\pm 0.35}}$ & $85.06_{\pm 0.75}$ \\
     CHAT       & $78.91_{\pm 0.16}$  & $84.11_{\pm 0.81}$ & $85.01_{\pm 0.42}$ & $\texttt{NaN}$ & $\mathbf{85.94_{\pm 0.18}}$ & $85.72_{\pm 0.29}$ \\
    HOMEPAP    & $69.43_{\pm 0.08}$        & $70.37_{\pm 1.90}$ & $72.56_{\pm 2.35}$ & $71.26_{\pm 1.93}$ & $73.14_{\pm 2.09}$ & $\mathbf{73.53_{\pm 2.00}}$ \\
     MASS       & $79.85_{\pm 1.27}$        & $77.40_{\pm 2.18}$ & $81.12_{\pm 2.27}$ & $79.06_{\pm 1.89}$ & $\mathbf{84.09_{\pm 0.85}}$ & $82.49_{\pm 1.22}$ \\
    PhysioNet  & $75.73_{\pm 0.38}$        & $77.19_{\pm 0.94}$ & $78.97_{\pm 0.43}$ & $77.98_{\pm 0.89}$ & $\mathbf{79.82_{\pm 1.63}}$ & $78.83_{\pm 1.60}$ \\
    SOF        & $78.74_{\pm 1.81}$        & $82.61_{\pm 0.35}$ & $83.39_{\pm 0.67}$ & $83.70_{\pm 1.01}$ & $\mathbf{84.69_{\pm 0.73}}$ & $84.31_{\pm 0.57}$ \\
    \bottomrule
    \end{tabular}
    }
    \label{tab:sleep_results}
\end{table}

\section{Conclusion / Open Challenges}
While promising, the current findings on Mantis's superior performance for EEG analysis suggest a significant step forward in applying foundation models to neural data. Its ability to outperform a specialized EEG model highlights the potential of generalist architectures. Future work should focus on extending these experiments to include a wider range of BCI datasets and new tasks, like emotion recognition, to fully validate Mantis's generalizability. Furthermore, addressing the challenge of zero-shot learning on EEG data, perhaps through specialized normalization techniques such as Monge alignment~\cite{gnassounou2025psdnorm}, could unlock new avenues for leveraging these powerful models without extensive training.

\bibliographystyle{apalike}
\bibliography{bibliography}

\newpage


\appendix

\section{Architectures}
\subsection{CBraMod}
\label{sec:cbramod}

Given an EEG sample $\mbf{x} \in \mathbb{R}^{C \times T}$, where $C$ is the number of channels and $T$ is the number of timestamps, CBraMod first partitions the time axis into non-overlapping windows of length $t$ ($t=200$ used in pretraining), producing
\[
\mbf{x} \mapsto \mbf{X} \in \mathbb{R}^{C \times p \times t}, 
\quad p = \left\lfloor \frac{T}{t} \right\rfloor.
\]

Each patch $\mbf{x}_{i,j}$ (short time series from channel $i$ and window $j$) is independently encoded via:
\begin{itemize}
    \item a time-domain convolutional branch $f_{\mathrm{conv}}$ (3-layer 1D CNN),
    \item a frequency-domain branch $W_{\mathrm{fft}} \cdot \mathrm{FFT}(\cdot)$ (FFT + linear projection).
\end{itemize}
Formally,
\[
\mbf{e}^t_{i,j} = f_{\mathrm{conv}}(\mbf{x}_{i,j}) \in \mathbb{R}^{200}, \quad
\mbf{e}^f_{i,j} = W_{\mathrm{fft}} \cdot \mathrm{FFT}(\mbf{x}_{i,j}) \in \mathbb{R}^{200}.
\]

The embeddings are combined as
\[
\mbf{e}_{i,j} = \mbf{e}^t_{i,j} + \mbf{e}^f_{i,j}, 
\quad \mbf{E} \in \mathbb{R}^{C \times p \times 200}.
\]

Asymmetric Conditional Positional Encoding (ACPE) generates spatial-temporal offsets $\mbf{e}^{pos}_{i,j} \in \mathbb{R}^{200}$, which are added to the patch embeddings:
\[
\mbf{o}_{i,j} = \mbf{e}_{i,j} + \mbf{e}^{pos}_{i,j}, 
\quad \mbf{O} \in \mathbb{R}^{C \times p \times 200}.
\]

The embeddings $\mbf{O}$ are processed through $L=12$ criss-cross transformer blocks, each with parallel spatial and temporal attention:
\begin{align*}
\mbf{O}_j &\in \mathbb{R}^{C \times 200}, &
\mathrm{S\text{-}Attn}(\mbf{O}_j) &= \mathrm{Attention}(\mbf{O}_j W^Q, \mbf{O}_j W^K, \mbf{O}_j W^V), \\
\mbf{O}_i &\in \mathbb{R}^{p \times 200}, &
\mathrm{T\text{-}Attn}(\mbf{O}_i) &= \mathrm{Attention}(\mbf{O}_i W^Q, \mbf{O}_i W^K, \mbf{O}_i W^V).
\end{align*}

Both attentions use $K=8$ heads with hidden dimension $d=200$, and the concatenated outputs yield
\[
\mbf{E}_r \in \mathbb{R}^{C \times p \times 200}.
\]

\paragraph{Pretraining}
For masked autoencoding, each representation is projected back to the time domain:
\[
\hat{\mbf{x}}_{i,j} = W_r \mbf{e}^r_{i,j} \in \mathbb{R}^{t}, 
\quad \hat{\mbf{X}} \in \mathbb{R}^{C \times p \times t}.
\]

When pretraining, $50\%$ of the patches are masked. The reconstruction loss is applied only to masked patches:
\[
\mathcal{L}_{\mathrm{MAE}} = \|\hat{\mbf{X}}_M - \mbf{X}_M \|_2^2.
\]
\subsection{Mantis}
\label{sec:mantis}

Given a time series sample $\mbf{x} \in \mathbb{R}^{C \times T}$, Mantis first
resizes each channel to a fixed length $t$ multiple of $32$ ($t=512$ used in pretraining) via interpolation and applies
instance-level standardization (per-channel mean and variance over time).  
For channel $i$, let $\mbf{x}^{(i)} \in \mathbb{R}^{t}$ denote the normalized series.

For each channel, a single 1D convolution layer (output width $256$) followed by mean pooling produces $32$ base patches.  
The same pipeline applied to the first difference $\Delta \mbf{x}^{(i)}$ yields $32$ differential patches.  
From the \emph{raw} (pre-normalization) series, per-patch statistics (mean $\mu_j$ and standard deviation $\sigma_j$) are computed and encoded via a scalar encoder.  
Concatenating the three feature parts and projecting yields the final tokens:
\[
\begin{aligned}
\mbf{c}_j &= \mathrm{MeanPool}\big(\mathrm{Conv}(\mbf{x}^{(i)})\big)_j \in \mathbb{R}^{256}, \\
\mbf{c}^{\Delta}_j &= \mathrm{MeanPool}\big(\mathrm{Conv}(\Delta \mbf{x}^{(i)})\big)_j \in \mathbb{R}^{256}, \\
\mbf{s}_j &= \mathrm{ScalarEnc}(\mu_j, \sigma_j) \in \mathbb{R}^{64}, \\
\mbf{t}_j &= \mathrm{LayerNorm}\!\Big(W_{\mathrm{proj}} \,[\,\mbf{c}_j ; \mbf{c}^{\Delta}_j ; \mbf{s}_j\,]\Big) \in \mathbb{R}^{256}, 
\quad j = 1, \dots, 32,
\end{aligned}
\]
so that 
\[
T^{(i)} = [\mbf{t}_1, \dots, \mbf{t}_{32}] \in \mathbb{R}^{32 \times 256}.
\]

A learnable class token $\mbf{t}_{\mathrm{cls}}$ is prepended, and sinusoidal positional encodings $P$ are added:
\[
\tilde{T}^{(i)} = [\mbf{t}_{\mathrm{cls}}; T^{(i)}] + P \in \mathbb{R}^{33 \times 256}.
\]

The sequence $\tilde{T}^{(i)}$ is processed through $L=6$ Transformer encoder blocks (each with $H=8$ heads; dropout $0.1$ during pretraining), and the class embedding is taken as the channel descriptor:
\[
\mbf{z}^{(i)} = \mathrm{ViT}_L(\tilde{T}^{(i)})_{\mathrm{cls}} \in \mathbb{R}^{256}.
\]

All channels are encoded independently and concatenated:
\[
\mbf{z} = \mathrm{concat}\big(\mbf{z}^{(1)}, \dots, \mbf{z}^{(C)}\big) \in \mathbb{R}^{256 \cdot C}.
\]

\paragraph{Pretraining} 
Mantis was pretrained using a contrastive loss.  
Let $\mbf{x}$ and $\mbf{x}'$ be two augmented views of the same original time series, and let their encoded representations be
\[
\mbf{z} = \mathrm{Mantis}(\mbf{x}), \qquad
\mbf{z}' = \mathrm{Mantis}(\mbf{x}') \in \mathbb{R}^{256 \cdot C}.
\]

We define the cosine similarity between two vectors as
\[
\mathrm{scos}(\mbf{a}, \mbf{b}) = \big\langle \frac{\mbf{a}}{\|\mbf{a}\|_2}, \frac{\mbf{b}}{\|\mbf{b}\|_2} \big\rangle.
\]

For a batch of $N$ samples, the contrastive (InfoNCE) loss for the $i$-th sample is
\[
\mathcal{L}_i = - \log \frac{\exp\big(\mathrm{scos}(\mbf{z}_i, \mbf{z}'_i) / \tau\big)}
{\sum_{j=1}^{N} \exp\big(\mathrm{scos}(\mbf{z}_i, \mbf{z}'_j) / \tau\big)},
\]
where $\tau > 0$ is a temperature hyperparameter.  
The total loss is averaged over the batch:
\[
\mathcal{L} = \frac{1}{N} \sum_{i=1}^{N} \mathcal{L}_i.
\]

This training encourages embeddings of augmented views of the same sample to be close, while pushing apart embeddings of different samples, yielding representations that capture meaningful temporal features invariant to augmentations.

\section{Experimental Setup}
\begin{table}[h]
    \centering
    \caption{Datasets and number of subjects for sleep datasets.}
    \label{tab:datasets_subjects}
    \resizebox{0.75\linewidth}{!}{
        \begin{tabular}{c|c c c c c c c c c }
        \toprule
            Dataset & ABC & CCSHS & CFS & HPAP  & PHYS  & MASS & CHAT & SOF \\
            \midrule
            Subjects & 44 & 515 & 681 & 166  & 70 & 61 & 1230 & 434 \\
            \bottomrule
        \end{tabular}
    }
    \label{tab:dataset}
\end{table}
Table~\ref{tab:dataset} reports the number of subjects in the sleep datasets. As shown, the number of subjects varies widely, ranging from 44 in ABC to 1,230 in CHAT.
For sleep dataset, we adopt a standard pre-processing step commonly used in sleep staging studies~\citep{chambon2018deep,Stephansen_2018}.
To ensure consistency across, we restrict the analysis to two bipolar EEG channels.
For the NSRR datasets, we select C3-A2 and C4-A1, while for Physionet and MASS, only Fpz-Cz and Pz-Oz are available and thus used.
All EEG signals are low-pass filtered at 30 Hz and resampled to 100 Hz. For CBraMod, the data are resampled to 200Hz and split into 1s patches.
Data extraction and preprocessing are performed with MNE-BIDS~\citep{Appelhoff2019} and MNE-Python~\citep{GramfortEtAl2013a}.

\end{document}